\documentclass{article}

\usepackage{arxiv}

\usepackage[utf8]{inputenc}
\usepackage[T1]{fontenc} 
\usepackage{hyperref}     
\usepackage{url}          
\usepackage{booktabs}      
\usepackage{amsfonts}     
\usepackage{nicefrac}      
\usepackage{microtype}     
\usepackage{multirow}
\usepackage{graphicx}
\usepackage{lipsum}
\usepackage{float}
\usepackage{natbib}

\usepackage{tabularx,booktabs,ragged2e}
\usepackage{rotating}                      % landscape tables
\usepackage{newtxmath}                     % italic greek letters
\usepackage[inline]{enumitem}              % inline enumerations
\usepackage{caption,subcaption}            % figure and table captions
\usepackage{eurosym}                       % euro symbol
\usepackage{xcolor}                        % color-coding

\title{Fairness in Credit Scoring: Assessment, Implementation and Profit Implications}

\author{
  Nikita Kozodoi \\
  Humboldt University of Berlin \\
  Berlin, Germany \\
  \And
  Johannes Jacob \\
  Humboldt University of Berlin \\
  Berlin, Germany \\
  \And
  Stefan Lessmann \\
  Humboldt University of Berlin \\
  Berlin, Germany \\
}

\begin{document}

\maketitle

\begin{abstract}
The rise of algorithmic decision-making has spawned much research on fair machine learning (ML). Financial institutions use ML for building risk scorecards that support a range of credit-related decisions. Yet, the literature on fair ML in credit scoring is scarce. The paper makes three contributions. First, we revisit statistical fairness criteria and examine their adequacy for credit scoring. Second, we catalog algorithmic options for incorporating fairness goals in the ML model development pipeline. Last, we empirically compare different fairness processors in a profit-oriented credit scoring context using real-world data. The empirical results substantiate the evaluation of fairness measures, identify suitable options to implement fair credit scoring, and clarify the profit-fairness trade-off in lending decisions. We find that multiple fairness criteria can be approximately satisfied at once and recommend separation as a proper criterion for measuring the fairness of a scorecard. We also find fair in-processors to deliver a good balance between profit and fairness and show that algorithmic discrimination can be reduced to a reasonable level at a relatively low cost. The codes corresponding to the paper are available on GitHub\footnote{The code is available at: \url{https://github.com/kozodoi/Fair_Credit_Scoring}}.
\end{abstract}

% keywords can be removed

%%%%%%%%%%%%%%%%%%%%%%%%%%%%%%%%%%%%
%                                  
%             MAIN BODY
%                                  
%%%%%%%%%%%%%%%%%%%%%%%%%%%%%%%%%%%%

\section{Introduction} 
\label{sec_introduction}

% scoring models
Financial institutions increasingly rely on machine learning (ML) to support decision-making \citep{crook2007recent}. The paper considers ML applications in the retail credit market, which is a large and economically important segment of the credit industry. For example, the total outstanding amount of retail credit in the US exceeded \$4,161 billion in 2020\footnote{Source: \url{https://www.federalreserve.gov/releases/g19/current}}. ML-based scoring models, also called scorecards, have played a major role in the approval of the corresponding loans.

% political concerns
In 2016, the Executive Office of the President of the US published a report on algorithmic systems, opportunity, and civil rights \citep{WhiteHouseReport}, which highlights the dangers of automated decision-making to the detriment of historically disadvantaged groups. It emphasizes credit scoring as a critical sector with a large societal impact, calling practitioners for using the principle of ``equal opportunity by design" across different demographic groups. Similar actions were taken by the EU when they supplemented their General Data Protection Regulation with a guideline that stresses the need for regular and systemic monitoring of the credit scoring sector \citep{DPO}. The guidelines issued by the EU and the US evidence political concern that potential violations of anti-discrimination law in credit scoring might affect debt and wealth distributions and have undesired economic effects on the society \citep{RN1}.

% fair ML
A growing literature on fair ML echos these concerns and proposes a range of statistical fairness measures and approaches for their optimization. It is common practice to discuss algorithmic fairness through the lens of differences between groups of individuals. The groups emerge from one or multiple categorical attributes that are considered sensitive. Examples include gender, religious denomination or ethnic group. The goal of fair ML is then to ensure that model predictions meet statistical fairness criteria. \cite{RN46} distinguishes 21 such criteria while \cite{RN7} show that most criteria can be derived from one of three main fairness measures: independence, separation, and sufficiency. Beyond quantifying fairness in  model-based predictions, fairness criteria also serve as constraints or objectives in the optimization problem that underlines the training of an ML model. Approaches to adjust model training to optimize fairness criteria next to common indicators of model fit are known as fairness processors.

% fair credit scoring
Surprisingly, the literature on fair ML and credit scoring share few touching points. As we detail in Section \ref{sec_related_work}, only three studies \citep{RN19, RN18, RN1} have considered the interface between the two disciplines. None of them focuses on operational decisions in the loan approval process and the potential trade-off between fairness and profit. Therefore, the goal of the paper is to i) provide a broad overview and systematization of recently developed fairness criteria and fairness processors, and to ii) empirically test their adequacy for credit scoring. While the fairness enhancing procedures that we consider are not new and have been developed in the fair ML literature, we suggest that our holistic and integrative perspective is useful to help risk analysts stay abreast of recent developments in that literature, judge their impact on credit scoring practices, and focus future research initiatives concerning fair credit scoring.

% contribution I
In pursuing its objective, the paper makes the following contributions: First, we revisit statistical criteria for measuring fairness and examine whether these criteria and their underlying understanding of distributional equality are appropriate for credit scoring. Given that different fairness criteria typically conflict with one another \citep{RN20}, our analysis is useful to inform the selection of a suitable fairness criterion (or set of criteria). Considering the relative costs of classification errors for banks and retail clients, we identify separation as a preferable criterion to appraise fairness in a lending context. More generally, our analysis may raise awareness for the risk of algorithmic discrimination in credit scoring, which, given the sparsity of prior work on the topic, may be seen as a valuable contribution to the credit risk community.

% contribution II
Second, we review and catalog state-of-the-art fairness processors across multiple important dimensions, including the target fairness criterion, the implementation method, and requirements for the classification problem. The catalog provides a systematic overview of fairness processors and clarifies whether and when these meet requirements associated with loan approval processes and the application context of credit scoring. The catalog also addresses the critique of \cite{RN8}, who demand a more uniform fairness terminology among scholars.

% contribution III
Last, we empirically compare a range of different fairness processors along several performance criteria using seven real-world credit scoring data sets. Unlike prior studies on fair ML, our analysis recognizes prediction performance indicators that are established in credit scoring and, importantly, the profitability of a scoring model. Furthermore, to extend the conceptual discussion on the suitability of the fairness criteria for credit scoring, we measure fairness not only with the criterion optimized by a processor but a range of different fairness criteria. The corresponding results provide original insights concerning the agreement among fairness criteria in credit scoring and their compatibility with profit. More specifically, our comparative analysis contributes to the empirical credit scoring literature by identifying fairness processors that best serve the interests and requirements of risk analysts and by elucidating the trade-off between profitability and fairness of a credit scoring system. A deeper understanding of this trade-off is crucial for managers and policy-makers to decide on the deployment of fairness enhancing procedures in financial institutions and regulatory directives to enforce certain levels of fairness, respectively.

\section{Theoretical Background} 
\label{sec_2}

% summary
This section covers relevant background on fair ML. We first examine methods to integrate fairness constraints into the model development pipeline and than review established fairness criteria. We focus on independence, separation and sufficiency because these criteria encompass a variety of other fairness concepts \citep{RN7, RN8}. Table A1 in the online Appendix details how independence, separation and sufficiency have synonymously been referred to in the literature and how they relate to the other formulations of fairness.

%%%%%%%%%%%%%%%%%%%%%%%%%%%%
%% FAIRNESS PROCESSORS
%%%%%%%%%%%%%%%%%%%%%%%%%%%%

\subsection{Fairness Optimization in the Modeling Pipeline}

% intro
Research on fair ML has recently emerged from the continuous integration of automated decision-making into important areas of social life and fairness concerns arising during this process \citep{RN59}. Much fair ML literature focuses on classification settings in which an unprivileged demographic group experiences discrimination through a classification model \citep{RN8}. Several attempts have been made to formalize the concept of fairness. Incorporating the corresponding fairness criteria in the ML pipeline facilitates measuring the degree to which class predictions discriminate against minorities \citep{RN7}. 

\begin{figure}[b]
	\centering
    \includegraphics[width = \textwidth]{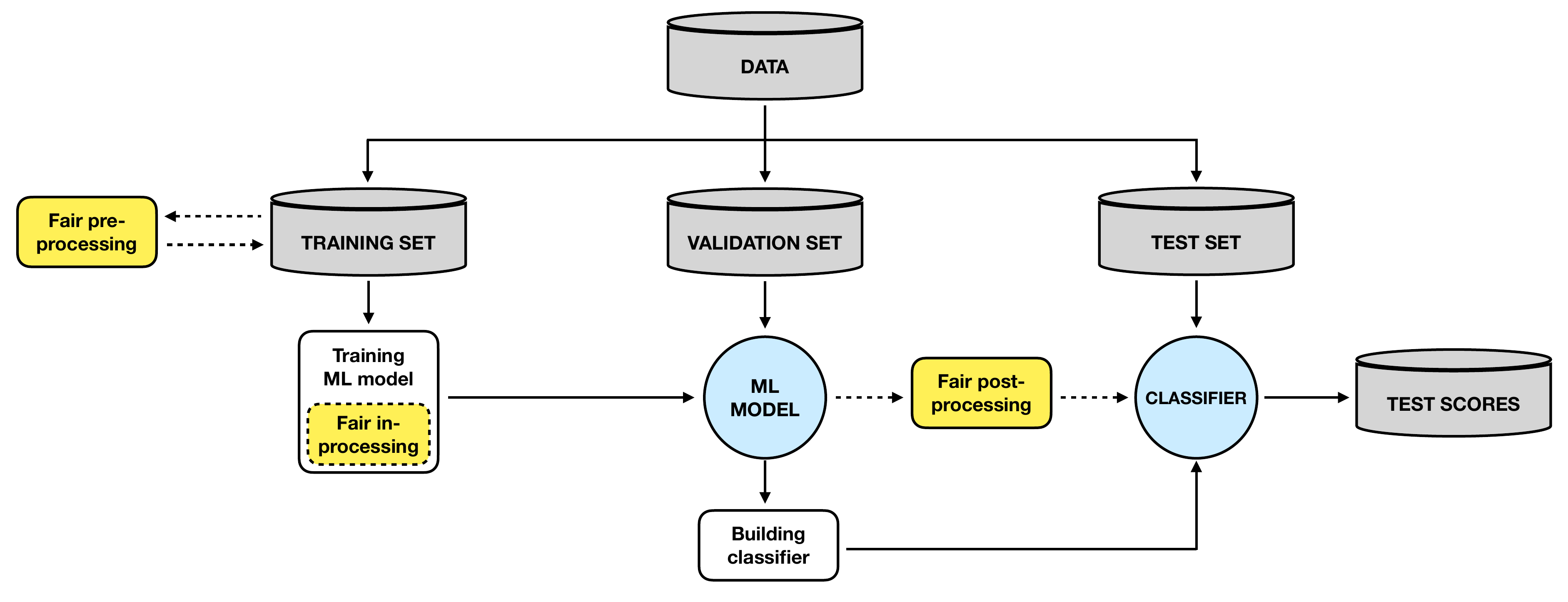}
	\caption{Fairness Integration in the ML Pipeline: In-processing, Pre-processing and Post-processing.}
	\label{fig_pipeline}
\end{figure}

% processor types
Algorithmic interventions designed to implement statistical fairness constraints are denoted as fairness processors. A processor can alter different stages in the ML pipeline. The literature distinguishes three methods of intervention: pre-processing, in-processing and post-processing \citep{RN43}. Their application generally depends on the conceptual and technical feasibility of a given prediction task. Figure \ref{fig_pipeline} illustrates the fairness processors within an ML pipeline. We describe selected approaches from each group in Section \ref{sec_4}.

% fairness pipeline

% pre-processing
Integrating a fairness processor into the pre-processing stage transforms the training data such that the input to a model is fair with respect to one or more sensitive features. Typically, fair pre-processing involves decorrelating the feature space with the sensitive attribute \citep[e.g.,][]{RN29}. Even though modifying the training data is sometimes not possible or practical, the advantage of fair pre-processing is that if fairness is ensured before ML model training, it will also be ensured during the next model development steps \citep{RN7}.

% in-processing
In-processing methods introduce auxiliary fairness constraints during ML model training. Then, training involves minimizing the empirical risk of the model while also optimizing a fairness criterion. In-processing renders a learned classifier (approximately) fair for the training data \citep{RN30}. Optimizing fairness during training has the potential to generate the highest utility as the tuning process also considers the fairness constraint. At the same time, in-processors are typically developed for settings with specific requirements (e.g., supporting only a single sensitive attribute), which limits their generality \citep{RN7}. Another disadvantage is that implementing a fair in-processor requires full access to the training process and the input data. This is especially problematic in heavily regulated domains such as credit scoring, where changes to a risk model might require regulatory approval and are associated with high costs.

% post-processing
After an ML model is trained, post-processing can be applied to adjust the learned classifier or change its predictions according to the requirements of a particular fairness criterion \citep{RN18}. The standard procedures include modifying the predicted scores or labels for specific observations. Unlike pre- or in-processing, post-processors need no information about the input data or the base model. This has the advantage that post-processors can be applied to any set of predictions. However, generality has a price. Post-processing is often less effective than alternative approaches and may substantially decrease classification accuracy \citep{RN7}.

%%%%%%%%%%%%%%%%%%%%%%%%%%%%
%% FAIRNESS CRITERIA
%%%%%%%%%%%%%%%%%%%%%%%%%%%%

\subsection{Fairness Criteria} 

% intro
This subsection introduces three established fairness criteria from a credit scoring perspective. Consider a setting in which a financial institution uses data on previous customers to predict whether a loan applicant will default. Let $X \in \mathbb{R}^k$ denote the $k$ features of a loan applicant and $y \in \{0, 1\}$ a random variable indicating if the applicant repays the loan ($y = 1$) or defaults ($y = 0$). The institution approves applications using a scoring model that predicts risk scores $s(X) = \mathbf{P}(y = 1|X)$. The score function can be turned into a classifier by accepting customers with scores above a cutoff $\tau$. Let $x_a  \in \{0, 1\}$ denote a protected attribute associated with certain characteristics of an applicant. For example, $x_a$ could indicate whether she has a disability ($x_a = 1$) or not ($x_a = 0$). Clearly, the value of $x_a$ must not impact the decision of the credit institution.

% note
In the following, we consider a binary protected attribute to simplify the exposition. The discussed fairness criteria generalize to multinomial protected attributes (i.e., protected attributes with more than two unique values). Also, note that the fair ML literature often uses the terms protected attribute and sensitive attribute interchangeably. From a methodological perspective, it is less important whether the use of an attribute is socially undesirable or regulated by law. We use the term sensitive attribute throughout the paper while acknowledging that our example attribute disability is not only sensitive but protected. 
The groups created when splitting individuals by a sensitive attribute are referred to as sensitive groups.

%%% INDEPENDENCE
\subsubsection{Independence}

The score $s(X)$ satisfies independence at a cutoff $\tau$ if the fraction of customers classified as good risks ($y = 1$) is the same in each sensitive group. Formally, this condition can be written as:
\begin{equation}
\label{eq_ind1}
\mathbf{P}\left[s(X\,|\,x_a = 0) > \tau\right] = \mathbf{P}\left[s(X\,|\,x_a = 1) > \tau\right]
\end{equation}
Equation (\ref{eq_ind1}) states that $s(X)$ is statistically independent of the sensitive attribute $x_a$ \citep{RN7}. Classifier predictions are not affected by the sensitive attribute, and the probability to be classified as a good risk is the same in both groups \citep{RN25}. In the prior work, the independence condition is also known as demographic or statistical parity \citep{RN20}.

This strict constraint is usually not feasible for real-world applications like credit scoring, as the resulting loss in model performance can make a business unsustainable. Therefore, it is a common practice in anti-discrimination law to allow the score and the sensitive attribute to share at least some mutual information and introduce a relaxation of the independence criterion \citep{RN59}. The Equal Opportunity Credit Act has a regulation that is referred to as the ``80 percent rule'' \citep{RN10}. The rule requires that $\mathbf{P}(s(X\,|\,x_a = 1) > \tau) \le 0.8\cdot \mathbf{P}(s(X\,|\,x_a = 0) > \tau)$, where $\{x_a = 0\}$ is the privileged group \citep{RN21}. 

Following the relaxation of the independence condition suggested in the prior work \citep{RN7}, we measure independence using a metric denoted as $\mbox{IND}$, which we define as: 
\begin{equation}
\label{eq_ind2}
\mbox{IND} = \left\lvert \mathbf{P}\left[s(X\,|\,x_a = 0) > \tau\right] - \mathbf{P}\left[s(X\,|\,x_a = 1) > \tau\right] \right\rvert
\end{equation}
A positive difference between the two terms implies that the group $\{x_a = 0\}$ is considered the privileged group and vice versa. The closer $\mbox{IND}$ is to zero, the lower is the discrimination.

%%% SEPARATION
\subsubsection{Separation}

The separation condition, also known as the equalized odds condition, is satisfied if the classification based on the predicted score $s(X)$ and the cutoff $\tau$ is independent on $x_a$ conditional on the true outcome $y$ \citep{RN7}. Formally, the score $s(X)$ satisfies separation at a cutoff $\tau$ if:
\begin{equation}
\label{eq_sp1}
\left\{
\begin{aligned}
\mathbf{P}\left[s(X\,|\,y = 0, x_a = 0) > \tau\right]   & = \mathbf{P}\left[s(X\,|\,y = 0, x_a = 1) > \tau\right] \\
\mathbf{P}\left[s(X\,|\,y = 1, x_a = 0) \le \tau\right] & = \mathbf{P}\left[s(X\,|\,y = 1, x_a = 1) \le \tau\right]
\end{aligned}
\right.
\end{equation}
The expression in the first line compares the false positive rate (FPR) across the sensitive groups, whereas the second line compares the false negative rate (FNR) per group. The separation criterion, therefore, requires that the FNR and the FPR are the same for the sensitive groups.

Separation acknowledges that $x_a$ may be correlated with $y$ (e.g., applicants with a disability might can a higher default rate). However, the criterion prohibits the use of $x_a$ as a direct predictor for $y$. When the difference between group sizes is large, the criterion will punish models that perform well only on the majority group \citep{RN18}. To measure the degree to which the separation condition is satisfied, we suggest using a criterion denoted as $\mbox{SP}$, which we define as:

\begin{equation}
\label{eq_sp2}
\mbox{SP} = \frac{1}{2} \left\lvert (\mbox{FPR}_{\{x_a = 1\}} - \mbox{FPR}_{\{x_a = 0\}}) + (\mbox{FNR}_{\{x_a = 0\}} - \mbox{FNR}_{\{x_a = 1\}}) \right\rvert
\end{equation}

$\mbox{SP}$ calculates the average absolute difference between the group-wise FPR and FNR.
A positive difference between each of the two group-wise error rates indicates that one of the groups has a lower misclassification rate. Perfect separation (i.e., $\mbox{SP} = 0$) is observed when the group-wise FPR and FNR are equal. Note that this also implies that the group-wise FPR and TPR are equal. Higher values of $\mbox{SP}$ indicate stronger discrimination through a larger difference in model performance across the sensitive groups.

%%% SUFFICIENCY
\subsubsection{Sufficiency}

The score $s(X)$ is sufficient at a cutoff $\tau$ if the likelihood that an individual belonging to a positive class is classified as positive is the same for both sensitive groups \citep{RN7}. This implies that for all values of $s(X)$ the following condition holds:
\begin{equation}
\label{eq_sf1}
\mathbf{P}(y = 1 \,|\, s(X) > \tau, x_a = 0) = \mathbf{P}(y = 1 \,|\, s(X) > \tau, x_a = 1)
\end{equation}
Equation (\ref{eq_sf1}) requires that the positive predictive value (PPV) is the same for the sensitive groups \citep{RN20}. This paper defines the sufficiency metric $\mbox{SF}$ as the absolute difference between the group-wise PPV:
\begin{equation}
\label{eq_sf2}
\mbox{SF} = \left\lvert \mbox{PPV}_{\{x_a = 0\}} - \mbox{PPV}_{\{x_a = 1\}} \right\rvert
\end{equation}
A large difference between the group-wise PPV indicates inconsistent model performance across the sensitive groups. The closer SF is to zero, the higher is the achieved sufficiency.

\section{Fairness and Credit Scoring} 
\label{sec_3}

% summary
The section discusses the interplay between fair ML and credit scoring. We summarize previous work in the field and examine the adequacy of fairness criteria for credit scoring.

%%%%%%%%%%%%%%%%%%%%%%%%%%%%
%% PREVIOUS WORK
%%%%%%%%%%%%%%%%%%%%%%%%%%%%

\subsection{Prior Work on Fair Credit Scoring}
\label{sec_related_work}

Prior literature on fair ML for credit scoring is surprisingly sparse. To our best knowledge, only three studies address algorithmic discrimination in credit scoring, and their focus differs substantially from that of this study. A first study by \cite{RN19} considers the credit market. The authors formalize the introduction of ML as a market intervention and examine the corresponding effect on interest rates in demographically different groups. \cite{RN1} take a similar perspective. Referring to the sample-selection bias, which arises from training scorecards on previously accepted cases \citep{RN47}, they argue that selection bias leads to scorecards overestimating the creditworthiness of some groups of applicants and perpetuates existing unfairness. To remedy this effect, \cite{RN1} call for mathematical constraints that optimize fairness as a long-term societal goal. However, the formulation of these constraints is still subject to further research. More generally, the long-term perspective of \cite{RN19} and \cite{RN1} emphasizes regulatory questions and is orthogonal to the focus on static fairness interventions, which prevails in the fair ML literature. These interventions address operational loan approval decisions and provide concrete approaches to remedy algorithmic bias.

Focusing on fairness interventions, a third study of \cite{RN18} is related to this paper more closely. \cite{RN18} propose the equalized odds fairness criterion and develop an algorithm that adjusts classifier predictions to raise fairness according to this criterion. The authors report enhanced fairness compared to a maximum profit benchmark using a credit scoring example based on FICO scores. In comparison to the focal paper, \cite{RN18} focus on the specific combination of one fairness processor and one fairness criterion. Their study does not examine the trade-off between profit and fairness and provides limited empirical evidence on how equalized odds compare to other fairness criteria or how fairness is best ensured in an ML pipeline. 

% summary
In summary, the main distinction between the focal paper and previous studies on fairness in credit scoring is that we undertake a comprehensive empirical analysis of alternative fairness criteria and fairness processors, which optimize these criteria. Prior work fails to account for the breadth of approaches that have been proposed in the scope of fair ML. Also, no previous study examines the interplay between fairness criteria and processors. Therefore, we aim at consolidating different advancements in fair ML, discussing their suitability for credit scoring, and providing rich empirical results that clarify the degree to which fairness constraints affect the predictive ability of credit scorecards and the corresponding profit implications, and how the trade-off between fairness and profit develops across fairness criteria and processors. We hope that our results offer actionable insights on how to set and pursue fairness objectives in credit scoring.

%%%%%%%%%%%%%%%%%%%%%%%%%%%%
%% FAIRNESS CRITERIA
%%%%%%%%%%%%%%%%%%%%%%%%%%%%

\subsection{Fairness Criteria for Credit Scoring}

% intro
The choice of the fairness criterion has severe consequences for the social impact of lending decisions \citep{RN1}. An unconstrained scoring model will take full advantage of the available (sensitive) information and discriminate between protected groups if this enhances predictive performance. The purpose of introducing fairness is, therefore, to adjust decision-making (i.e., scoring) practices for a better, discrimination-free outcome. According to the U.S. anti-discrimination law, for example, the demographic properties of a loan applicant should not influence lending decisions \citep{ECOA}. Arguably, the societal goal behind such law is an equal opportunity for financial well-being across demographically different groups. Achieving this goal in credit scoring is difficult as clients face unequal misclassification costs. Applicants that are denied a loan they could have repaid face the cost of a missed opportunity to enhance their social and economic position. However, if applicants receive a loan they cannot repay, they are confronted with financial debt and a long-term worsening of their financial situation as future access to financing will be more difficult. With these characteristics of  credit scoring in mind, the following considerations elaborate on the extent to which independence, separation and sufficiency fulfill the goal of equal opportunity for financial well-being in society.

% independence
Forcing independence on a scoring model's results in the same rate of accepted customers within sensitive groups. The problem with this approach is that the ability to repay a loan can have a different distribution in each group \citep{RN7}. If this is the case, but members of both groups have the same probability of receiving a loan, one group will experience more actual defaults. For a client, the consequences of defaulting can be more severe than the opportunity costs associated with a rejected application. Typically, the historically unprivileged group has a higher rate of non-solvent customers. Handing out loans to such individuals might worsen their financial situation in the long term \citep{RN18}. Instead of achieving fairness, this can lead to further perpetuating existing unfairness. The goal of better financial equality would not be met, and the financial gap in society could become even wider.

% separation
The separation criterion addresses this dilemma and acknowledges that a sensitive attribute might correlate with default rates. Requiring the same error rates between groups but allowing different positive classification rates, separation achieves a fair result that is closer to the reality of credit allocation decisions and more desirable from a customer's perspective. More precisely, separation accounts for different misclassification costs between groups. On the contrary, separation would be inadequate if credit scoring had a strictly preferred outcome for a customer, as is the case in domains like college admission \citep{RN8}. Interestingly, the first formulation of the separation criterion in the context of ML by \citep{RN18} is based on the example of the credit scoring domain and the limitations of the independence criterion to meet its requirements. 

% sufficiency
Sufficiency requires the ratio of true positive classifications over all positive classifications to be the same for the sensitive groups. This concept has two disadvantages for credit scoring. First, it allows for substantial discrimination in separation. For both groups, the proportion of correctly labeled non-default clients can be the same, satisfying sufficiency. In contrast, the likelihood of a potential non-default customer being classified as a bad risk can still differ between groups, violating the separation constraint. Second, most ML algorithms are designed to achieve sufficiency without integrating a fairness constraint if the model can predict the sensitive attribute from the other features \citep{RN7}. In credit scoring, the question would, therefore, be if the current procedure for assessing a customer's default risk and the associated distribution of loans is fair. The literature suggests a negative answer to this question \citep{RN19, RN1, RN18}. Hence, sufficiency appears less suitable for credit scoring. 

% conclusions I
Based on these considerations, the separation criterion appears most suitable to achieve a desirable form of fairness in credit scoring. Separation accounts for the imbalanced misclassification costs of the customer, and, as these imbalanced costs also exist for the financial institution, separation is also able to consider the interests of the loan market. 

% conclusions II
The considerations provided in this section suggest that the question of which fairness constraint is most adequate for credit scoring should be a part of a wider academic and societal debate. Such a democratic process should also acknowledge the importance of studying the long-term effects of implementing different fairness constraints to judge whether the societal goal of better financial equality between demographic groups can be achieved with specific interventions \citep{RN1}. 

\section{Methodology} 
\label{sec_4}

% outline
This section systematically reviews and catalogs fairness processors suggested in the prior work across different dimensions and discusses their applicability in credit scoring. Using the constructed catalog, we select and describe eight fairness processors that are part of the empirical study.

%%%%%%%%%%%%%%%%%%%%%%%%%%%%
%% CATALOG
%%%%%%%%%%%%%%%%%%%%%%%%%%%%

\subsection{Cataloging Fairness Processors}

% summary and dimensions
The fair ML literature has developed a variety of fairness processors to implement independence, separation and sufficiency constraints. The complexity between these processors varies considerably, from simply relabeling the prediction outcomes \citep[e.g.,][]{RN33} to complex deep learning approaches for training a discrimination-free classifier \citep[e.g.,][]{RN34}. Furthermore, some processors are limited to specific problem setups. This motivates us to develop a structured overview of fairness processors with respect to their characteristics and applicability. Specifically, we catalog existing fairness processors in Table \ref{tab_processors} using six dimensions: (i) point of intervention into the ML pipeline; (ii) optimized fairness criterion; (iii) classification problem type supported by a processor (binary or multinomial); (iv) possible number of sensitive attributes (one or multiple); and (vi) supported types of sensitive attributes (binary or multinomial).  

% processors catalog
\begin{sidewaystable}
    \centering
    \small
    \caption{Fairness Processors}
    \begin{tabular}{@{\extracolsep{3pt}} llccccccc}  
    \toprule
    Fairness processor & Reference & Method & Criterion & MT & MS & MA & PE & This paper \\  
    \midrule
    Reweighing & \cite{RN14}                            & PRE & IND  &  &  &  &  & $\checkmark$ \\  
    Massaging & \cite{RN14}                             & PRE & IND  &  &  &  &  &  \\  
    Classification without discrimination & \cite{RN13} & PRE & IND  &  &  &  &  &  \\  
    Discrimination discovery K-NN & \cite{RN50}         & PRE & IND  & $\checkmark$ &  &  &  &  \\  
    Fair representation learning & \cite{RN27}          & PRE & IND  & $\checkmark$ &  &  &  &  \\  
    Disparate impact remover & \cite{RN10}              & PRE & IND  &  & $\checkmark$ & $\checkmark$ &  & $\checkmark$ \\  
    Variational fair autoencoder & \cite{RN15}          & PRE & IND  & $\checkmark$ & $\checkmark$ & $\checkmark$ &  &  \\  
    Feature adjustment & \cite{RN3}                     & PRE & IND  & $\checkmark$ & $\checkmark$ & $\checkmark$ &  &  \\  
    Discrimination-free pre-processing & \cite{RN29}    & PRE & IND  &  & $\checkmark$ & $\checkmark$ &  &  \\  
    \midrule
    Prejudice remover regularizer & \cite{RN9}          & IN & IND         & $\checkmark$ &  &  &  & $\checkmark$ \\  
    Fair accuracy maximizer & \cite{RN16}               & IN & IND         & $\checkmark$ & $\checkmark$ & $\checkmark$ &  &  \\  
    Non-discriminatory Learner & \cite{RN45}            & IN & SP          &  &  &  &  &  \\  
    Adversarial debiasing & \cite{RN34}                 & IN & SP          & $\checkmark$ & $\checkmark$ & $\checkmark$ &  & $\checkmark$ \\  
    Meta-fairness algorithm & \cite{RN12}               & IN & IND, SP, SF &  & $\checkmark$ & $\checkmark$ &  & $\checkmark$ \\  
    \midrule
    Group-wise Platt scaling & \cite{RN35}, \cite{RN7}  & POST & SF  & $\checkmark$ & $\checkmark$ & $\checkmark$ &  & $\checkmark$ \\  
    Group-wise histogram binning & \cite{RN52}          & POST & SF  & $\checkmark$ & $\checkmark$ & $\checkmark$ &  &  \\  
    Group-wise isotonic regression & \cite{RN51}        & POST & SF  & $\checkmark$ & $\checkmark$ & $\checkmark$ &  &  \\
    Fairness-aware classifier & \cite{RN6}              & POST & IND &  &  &  &  &  \\  
    Reject option classification & \cite{RN33}          & POST & IND, SP &  & $\checkmark$ & $\checkmark$ &  & $\checkmark$ \\  
    Fairness constraint optimizer & \cite{RN49}         & POST & IND & $\checkmark$ & $\checkmark$ & $\checkmark$ &  &  \\  
    Equalized odds processor & \cite{RN18}              & POST & SP  &  & $\checkmark$ &  & $\checkmark$ & $\checkmark$ \\  
    Calibrated equalized odds & \cite{RN25}             & POST & SP  &  &  &  &  &  \\  
    \bottomrule
    \multicolumn{9}{l}{Abbreviations: IND = Independence, SP = separation, SF = sufficiency; PRE = pre-processor, IN = in-processor, POST = post-processor;} \\
    \multicolumn{9}{l}{MT = multinomial target, MS = multinomial sensitive attribute, MA = multiple sensitive attributes, PE = profit-driven evaluation.} \\
\end{tabular}
    \label{tab_processors}
\end{sidewaystable}

% conclusion I
Three main conclusions emerge from Table \ref{tab_processors}. First, the majority of processors implement the independence criterion. This may come the other criteria being invented only recently (see Table A1 in the online Appendix for comparison). Furthermore, independence allows implementation via pre-processing, which provides an additional point of intervention in the ML pipeline. In many scenarios, however, fairness through independence may not be a suitable choice. This calls for additional processors that implement the other two criteria. 

% conclusion II
Second, the choice of a suitable fairness processor is limited by the application and implementation context of a scorecard. The application context determines the type of target variable and sensitive attribute(s) to be handled by a processor. For instance, in a setup with multiple sensitive attributes optimizing separation is only possible via the adversarial debiasing or reject option classification. This is a severe limitation for credit scoring because financial institutions commonly face several protected attributes: the U.S. anti-discrimination law distinguishes nine bases that must not influence lending decisions, including race, color, religion and other customer attributes \citep{ECOA}. The implementation context can also limit possible points of intervention in the ML pipeline. Replacing a scorecard with a fair in-processor might require regulatory approval and incur additional costs. Post-processors are easier to implement since they are agnostic of the input data and the scorecard and only require access to the predicted scores.

% conclusion III
Third, it is a standard procedure to embed the fairness processor into an accuracy-optimizing framework. The loss in predictive accuracy is commonly used as a performance measure to judge the cost of integrating a fairness constraint. In line with this framework, \cite{RN11} conducted a comparative study to examine the achieved fairness and accuracy of four fairness processors. However, recent credit scoring literature criticizes the practice of using standard performance measures for evaluating scoring models and calls for profit-driven evaluation \citep{RN36}. In such a setup, evaluation of fairness processors should be performed with a profit maximization objective instead of standard statistical performance measures such as accuracy.

% summary
To conclude, the catalog suggests that a comparative analysis of fairness processors under profit maximization is needed to clarify the ``cost of fairness''. We argue that the profitability aspect is underrepresented in the fair ML literature, while it is highly relevant for real-world applications. A better understanding of the (dis)agreement of profitability and different fairness criteria is also useful for policy making as it sheds some light on the thorny question of which criterion lending institutions should emphasize. Which fairness processor to use for optimizing the desired criterion is yet another question with high relevance for practice. Prior literature offers limited guidance due to assessing processors typically only in terms of the single criterion that this processor implements. Contributing toward answering these pressing questions is the overall goal of the paper.

%%%%%%%%%%%%%%%%%%%%%%%%%%%%
%% PROCESSORS
%%%%%%%%%%%%%%%%%%%%%%%%%%%%

\subsection{Selected Fairness Processors}

% intro
This subsection overviews eight fairness processors from the catalog presented in Table \ref{tab_processors}. The selection of processors covers all combinations of fairness interventions. Following the setup introduced in Section \ref{sec_2}, we consider a credit scoring setup with a binary target variable $y \in \{0, 1\}$ and a binary sensitive attribute $x_a \in \{0, 1\}$ to introduce the processors. Some of the considered processors also generalize to multinomial target and sensitive attributes (see Table \ref{tab_processors} for details).

%%%%% PRE-PROCESSORS

\subsubsection{Pre-Processors}

%%% REWEIGHING

% idea and equation
Fairness pre-processors transform the input data to achieve fairness. Reweighing is a pre-processor that assigns weights to each observation in the training set based on the overall probabilities of the group-class combinations \citep{RN14}. Thus, weights for observations with $(x_a = 1, y = 1)$ are greater than weights for observations with $(x_a = 0, y = 1)$ if members of the group $\{x_a = 1\}$ have a lower probability to belong to a positive class than those of the group $\{x_a = 0\}$:
\begin{equation}
W(X\,|\,x_a = 1,\ y = 1) = \frac{\mathbf{P}_{exp}(x_a = 1\,|\,y = 1)}{\mathbf{P}_{obs}\left(x_a = 1\ \right|y = 1)}\ ,
\end{equation}
% explanation and example
where $\mathbf{P}_{exp}$ is the expected probability and $\mathbf{P}_{obs}$ is the observed probability. For instance, assume that $90\%$ of all individuals belong to the positive class and $20\%$ percent belong to the group $\{x_a = 1\}$. Then, $\mathbf{P}_{exp}(x_a = 1\,|\,y = 1) = 0.9\cdot 0.2=0.18$. If, in fact, only $12\%$ of all cases in $\{x_a = 1\}$ belong to the positive class, then $W\left(X\,|\,x_a = 1,y = 1\right) = \frac{0.18}{0.12}=0.9$. %The weights for the remaining combinations of $x_a$ and $y$ are calculated similarly.

% further explanation
Based on the computed weights, a fair training set is resampled with replacement such that combinations with a higher weight reappear more often. This procedure helps to fulfill the independence criterion. A discrimination-free classifier can then be trained on the resampled data.

%%% DI REMOVER

% idea and explanation
Another pre-processing technique is the disparate impact remover proposed by \cite{RN10}. The intuition behind this processor is to ensure independence by prohibiting the possibility of predicting the sensitive attribute $x_a$ with the other features in $X$ and the outcome $y$. This is achieved by transforming $X$ into $\overline{X}$ while preserving the rank of $X$ within sensitive groups defined by $x_a$. By preserving the rank of $X$ given $x_a$, the classification model $f(\overline{X})$ will still learn to choose higher-ranked credit applications over lower-ranked ones based on the other features.

% further explanation
The transformation is performed using an interpolation based on a quantile function and the cumulative distribution of $F:\mathrm{\mathbf{P}}(X\,|\,x_a = a)$. This ensures that given the transformed $\overline{X}$ at some rank, the probability of drawing an observation given $x_a = a$ is the same as for the entire data set. Hence, $x_a$ cannot be predicted with the other attributes, and the independence criterion is fulfilled. Since ensuring perfect independence can have a strong negative impact on a classifier utility, the transformation can be modified to only partially remove disparate impact. The meta-parameter $\lambda \in [0, 1]$ allows controlling the desired level of fairness-utility trade-off during transformation.

%%%%% IN-PROCESSORS

\subsubsection{In-Processors}

%%% PREJUDICE REMOVER

% idea and equation
In-processors achieve fairness when building a classifier. One of such methods, prejudice remover, introduces a fairness-driven regularization term to the classification model \citep{RN9}. Regularization is a standard statistical approach to penalize a model for some undesired behavior. This is typically done by adding a regularizer term to the loss function.

% PI index I
The fairness-driven regularization introduced by \cite{RN9} is based on the prejudice index $\mbox{PI}$, which quantifies the degree of unfairness based on the independence criterion:
\begin{equation}
\mbox{PI} = \sum_{(y, x_a) \in D}{\mathbf{P}(y,x_a) \ln \frac{\mathbf{P}(y,x_a)}{\mathbf{P}(x_a)\mathbf{P}(y)}} ,
\end{equation}
% PI index II
where $\mathbf{P}(y, x_a)$, $\mathbf{P}(y)$ and $\mathbf{P}(x_a)$ are empirical distributions of $y$ and $x_a$ over the sample $D$. $\mbox{PI}$ measures the amount of mutual information between $y$ and $x_a$. High values of $\mbox{PI}$ indicate that a sensitive attribute $x_a$ is a good predictor for $y$. The optimization problem extends to:
\begin{equation}
\mathop{\mathrm{min}}_{f} L\left[f\left(X\right),y\right] + \eta \mbox{PI}\ ,
\end{equation}
% further explanation
where $L(\cdot)$ is the underlying loss function of the model $f(X)$, and $\eta $ controls the importance of the term $\mbox{PI}$. In this study, we tune $\eta$ to maximize the profitability of a scorecard. The regularization term ensures that the sensitive attribute $x_a$ becomes less influential in the final prediction.

%%% ADVERSARIAL DEBIASING

% idea and explanation
Adversarial debiasing is another in-processor that stacks two neural networks with contrary objectives on top of each other \citep{RN34}. The first network (predictor) is trying to learn a function to predict $y$ given $X$, while also minimizing the success of the second network. The second network (adversary) takes the output layer of the first model $\hat{y}$ and the true labels $y$ as input and tries to predict the sensitive attribute $x_a$. Both models have objective-specific loss functions and weights that can be optimized using standard gradient-based optimization methods such as stochastic gradient descent or Adam \citep{adam}.

% some notations
The adversary is assumed to have weights $U$ and loss function $L_A\mathrm{(}\hat{x}_a,x_a\mathrm{)}$. The weights $U$ are updated according to the gradient ${\mathrm{\nabla }}_U L_A$ to minimize $L_A$. The weights of the predictor denoted as $W$ are modified based on a gradient that minimizes its loss function $L_P(\hat{y},y)$ but also maximizes the loss function of the adversary: ${\mathrm{\nabla }}_WL_P\left(\hat{y},y\right)-\ {\mathrm{\alpha }\mathrm{\ }\mathrm{\nabla }}_WL_A\left(\hat{x}_a,x_a\right)$, where $\alpha $ is a meta-parameter. 

% further explanation II
Since the adversary takes the output of the predictor $\hat{y}$ as input, the predictor aims to hold back any additional information about the sensitive attribute $x_a$ in its output $\hat{y}$ as it would improve the adversary's loss. In other words, the predictor will try to deceive the adversary and not share any additional information in $\hat{y}$. As $y$ is known to the adversary, the algorithm acknowledges that the sensitive attribute might correlate with $y$, and only unnecessary information will be avoided. Hence, the adversarially debiased model will converge towards the separation criterion.

%%% META-ALGORITHM

% idea
The meta fair classification algorithm is yet another in-processor designed to achieve fairness according to one of the different fairness criteria. For a given criterion, \cite{RN12} suggest using a corresponding group-wise fairness metric denoted as $\mbox{FM}$, where similar values of $\mbox{FM}$ across sensitive groups indicate a higher level of fairness. Given a classifier $f(X)$ with a loss function $L\left(f\left(X\right),y\right)$, they add a fairness constraint to the loss optimization problem during training:
\begin{equation}
    \label{eq_meta}
    \mathop{\mathrm{min}}_{f} L\left(f\left(X\right),y\right)
    \text{\quad s.t. \quad}
    \frac{\mathrm{min} \left[\mbox{FM}(f(X\,|\,x_a = 0)), \ \mbox{FM}(f(X\,|\,x_a = 1))\right]}{\mathrm{max} \left[\mbox{FM}(f(X\,|\,x_a = 0)), \ \mbox{FM}(f(X\,|\,x_a = 1))\right]}\  \ge\  \sigma ,
\end{equation}
% explanation I
where $\sigma \in [0,1]$ is a desired fairness bound. Higher values of the fraction in Equation \ref{eq_meta} indicate a higher similarity of $\mbox{FM}$ across sensitive groups, and $\sigma = 1$ implies perfect fairness.

% explanation II
For example, in case of sufficiency, FM is set to positive predictive value (PPV) such that $\mbox{FM}(f) = \mbox{PPV}(f) = \frac{\mathbf{P}(f=1\,|\,x_a=a,y=1)}{\mathbf{P}(f=1\,|\,x_a=a)}$. If the group $\{x_a = 1\}$ has a low PPV and the group $\{x_a = 0\}$ has a high PPV, the fraction in the optimization condition is close to zero. A high $\sigma$ will, therefore, bound the classifier to a high degree of fairness. During training, the value for $\sigma$ can be tuned such that it maximizes profit while minimizing the loss in fairness, i.e., the loss in sufficiency.

%%%%% POST-PROCESSORS

\subsection{Post-Processors}

%%% REJECT OPTION CLASSIFICATION

% idea I
As a post-processing method, reject option classification is based on the output of a learned classifier \citep{RN33}. In a credit scoring setup, the classifier output is a credit score that reflects the posterior probability to not default for each customer $s(X) = \mathbf{P} (\hat{y} = 1 | X)$. The closer the score is to $1$ or $0$, the higher is the certainty with which the classifier assigns the corresponding labels, whereas a score close to $0.5$ implies a high degree of uncertainty. 

% idea II
Reject option classification defines a critical region of high uncertainty and reassigns labels for customers that have predicted scores within this region, such that members of the unprivileged group receive a positive label ($y = 1$) and vice versa. Formally, the critical region is defined as:
\begin{equation}
\mathrm{max} \left[ \mathbf{P} (\hat{y} = 1 | X) \ , \ 1 - \mathbf{P} (\hat{y} = 1 | X) \right] \ \le \theta \ ,
\end{equation}
% equation explanation
where $0.5 < \theta < 1$. Given a set of predicted scores and the true outcomes, a suitable value of $\theta$ and the number of required posterior reclassifications can be tuned to optimize a fairness criterion (e.g., independence) within a specified interval restricted by the lower and the upper bound of the fairness metric denoted as $[\sigma_l, \sigma_u]$.

%%% EQUALIZED ODDS PROCESSOR

% idea
Equalized odds processor uses a different logic to post-process classifier predictions. It finds a cutoff value $\tau$ that optimizes the predictive performance while satisfying the separation criterion, i.e., ensuring the same false negative and false positive rate per group \citep{RN18}.

% explanation and equation
Consider the receiver operating characteristic (ROC) curves that depict the trade-off between true and false positive rates for two sensitive groups. In an unfair scenario, the group-wise ROC curves have different slopes, which implies that not all trade-offs are achievable in each group. In the accuracy optimization setting, the optimal cutoff that satisfies sufficiency lies at the intersection of group-wise ROC curves. When optimizing for profit, the misclassification costs are not the same for both error rates. Thus, the optimal cutoff could lie somewhere else. Given a loss function $L(\cdot)$, \citet{RN18} suggest to derive a suitable cutoff $\tau$ by optimizing the following objective:
\begin{equation}
\min \mathbf{P} \left(s(X | x_a = a, y = 0) \le \tau \right) \cdot L(\hat{y} = 1, y = 0) + [1 - \mathbf{P} \left(s(X | x_a = a, y = 1) > \tau \right)] \cdot L(\hat{y} = 0,y = 1)
\end{equation}

%%% PLATT SCALING

% calibration idea
Platt scaling is a post-processing method that stems from the notion of calibration \citep{RN35}. Calibration addresses the problem that some classification algorithms are not able to make a statement about the certainty of their prediction, i.e., the probability with which an instance belongs to a certain class. In credit scoring, the predicted score could be an indicator of default risk but not the actual probability of default. A score $s(X)$ is calibrated if $\mathbf{P} \left(y = 1\ \right|\ s(X) = \tau) = \tau$.

% extending to groups
When extending the calibration condition to the group level, it becomes apparent that it implements the sufficiency criterion (see \cite{RN7} for proof):
\begin{equation}
\mathbf{P} \left[y = 1\ |\ s(X) = \tau,\ x_a = 1\right] = \ \mathbf{P} \left[y = 1\ |\ s(X) = \tau,\ x_a = 0\right] = \tau
\end{equation}

% Platt scaling
To achieve calibration per group, Platt scaling is applied separately to each sensitive group. The method uses the output of a possibly uncalibrated score $s(X)$ as input for logistic regression fitted against the target variable $y$. Based on the loss function of the logistic regression, the result is a new calibrated score that represents the probability that an instance belongs to the positive class. Formally, Platt scaling minimizes the log-loss $-\mathbb{E}[y log\left(\sigma \right) + \left(1 - y\right)\mathrm{log}\mathrm{}(1 - \sigma)]$ by finding the optimal parameters $a$ and $b$ of the sigmoid function $\sigma =\frac{1}{1 + \mathrm{exp}\mathrm{}(aS+b)}$.

\section{Experimental Setup}
\label{sec_5}

%%%%%%%%%%%%%%%%%%%%%%%%%%%%
%% DATA
%%%%%%%%%%%%%%%%%%%%%%%%%%%%

\subsection{Data}

% data sources
The empirical experiment is based on seven credit scoring data sets. Data sets \textit{german}\footnote{Source: \url{https://archive.ics.uci.edu/ml/datasets/statlog+(german+credit+data)}} and \textit{taiwan}\footnote{Source: \url{https://archive.ics.uci.edu/ml/datasets/default+of+credit+card+clients}} stem from the UCI Machine Learning Repository. Data sets \textit{Pakdd}\footnote{Source: \url{https://www.kdnuggets.com/2010/03/f-pakdd-2010-data-mining-competition.html}}, \textit{gmsc}\footnote{Source: \url{https://kaggle.com/c/givemesomecredit}} and \textit{homecredit}\footnote{Source: \url{https://kaggle.com/c/home-credit-default-risk}} were provided by different companies for the data mining competitions on PAKDD and Kaggle. Data sets \textit{Bene} and \textit{uk} were collected from financial institutions in the Benelux and UK \citep{lessmann2015benchmarking}.

% data summary
Each data set has a unique set of features describing a loan applicant and loan characteristics. The target variable $y$ is a binary indicator of whether the applicant has repaid the loan ($y = 1$) or not ($y = 0$). Each data set also contains a sensitive demographic attribute $x_a$ indicating the applicant's age group. The Equal Credit Opportunity Act prohibits that demographic characteristics such as the applicants' age impact credit approval decisions. We distinguish two groups of applicants: $\{x_a = 1\}$ contains applications where the applicant's age is below $\psi$ years, and $\{x_a = 0\}$ refers to the applications from customers older than $\psi$. We set $\psi = 25$, following the findings of \cite{RN13}, who used one of the consumer credit scoring data sets to discover that applicants from different age groups exhibit the greatest disparate impact (i.e., difference in $\mathbf{P}\left[y=1\, | \,x_a = a\right]$) at a threshold of $25$ years. Table \ref{tab_datasets} summarizes the main characteristics of the data sets.

% data summary
\begin{table}[t]
    \centering
    \small
    \caption{Credit Scoring Data Sets}
    \begin{tabular}{lcccc}
     \toprule
     Data set & Sample size & No. features & Default rate & Sensitive group rate \\ 
     \midrule
     german      & 1,000    & 61    & .30 & .19 \\
     bene        & 3,123    & 82    & .33 & .12 \\
     taiwan      & 23,531   & 76    & .23 & .14 \\
     uk          & 30,000   & 51    & .04 & .20 \\
     pakdd       & 50,000   & 185   & .26 & .11 \\
     gmsc        & 150,000  & 68    & .07 & .02 \\
     homecredit  & 307,511  & 92    & .08 & .04 \\
     \bottomrule
\end{tabular}
    \label{tab_datasets}
\end{table}

%%%%%%%%%%%%%%%%%%%%%%%%%%%%
%% SETUP
%%%%%%%%%%%%%%%%%%%%%%%%%%%%

\subsection{Experimental Setup}

% modeling pipeline
On each data set, we implement the eight fairness processors introduced in Section \ref{sec_4}, following the model development pipeline depicted in Figure \ref{fig_pipeline}\footnote{The code is available at: \url{https://github.com/kozodoi/Fair_Credit_Scoring}}. First, we partition the data into training ($60\%$) and test ($40\%$) sets. We then perform five-fold cross-validation on the training set. Each of the five combinations of training folds is used to train a scoring model and implement fairness processors. An unconstrained scoring model (i.e., a model that does not include any fairness-optimizing procedures) serves as a benchmark and represents the profit maximization scenario. Next, we consider in-processors in the form of the prejudice remover, adversarial debiasing and the meta fair algorithm. Relying on an in-processor implies that the trained in-processor serves as a scorecard. This contrasts pre- and post-processors, in which the actual scorecard is still based on a conventional ML algorithm. We consider reweighing and the disparate impact remover to pre-process (i.e., transform) the training data before developing a scoring model. Reject option classification, the equalized odds processor and Platt scaling represent the post-processors in our study. To learn a post-processing model, we apply each of them to the validation fold predictions of the unconstrained scorecard.

% base models
Fairness pre- and post-processors, as well as an unconstrained scorecard, use four base classifiers: logistic regression, artificial neural network and the tree-based ensemble learners random forest and extreme gradient boosting (XGB). Using multiple base learners allows us to check the robustness of processors across different classifiers. The base learners are established in credit scoring \citep[e.g.,][]{lessmann2015benchmarking, kozodoi2019multi}, whereby XGB \citep{chen2016xgboost} is maybe less known in the community. We include XGB due to its reputation as a highly powerful learning algorithm in Kaggle competitions and its strong performance in a recent credit scoring study by \cite{gunnarsson2021deep}, who find XGB outperforming challenging deep learning benchmarks.
Meta-parameters of the base classifiers are tuned in a nested four-fold cross-validation on the training data. The meta-parameters of fairness processors are also tuned using grid search. The details on the meta-parameter values and the tuning procedure are provided in the online Appendix. 

% profit I
Fairness processors and benchmarks are evaluated on the test set using multiple performance metrics. First, we measure the profitability of a scorecard by computing profit per EUR issued by a financial institution. To estimate profit, we start from the Expected Maximum Profit (EMP) criterion \citep{RN36}. The EMP measures the incremental profit compared to a base scenario in which loan applications are accepted without screening. This often leads to a small magnitude of EMP differences across classifiers \citep{kozodoi2019multi} and complicates the interpretation of the metric. To enable a more direct interpretation, we normalize misclassification costs such that the base scenario represents rejecting all applications.

% profit II
Table \ref{tab_costs} provides the confusion matrix of a scoring model, where $\pi_i$ are prior probabilities of good and bad risks, and $F_i(\tau)$ are predicted cumulative density functions of the scores of class $i$ given a cutoff value $\tau$. 
If an applicant is predicted to be a good risk, a financial institution faces cost $B$ in case of an incorrect prediction and earns $C$ from an accurate prediction. In contrast, if an applicant is predicted to be a bad risk, a company faces an opportunity cost $C$ in case of an incorrect prediction. Parameters $B$ and $C$ are defined according to \citet{RN36}.

% cost matrix for profit computation
\begin{table}[t]
    \centering
    \small
    \caption{Cost Matrix for Profit Computation}
    \begin{tabular}{@{\extracolsep{30pt}} lcc} 
	\toprule
	& \multicolumn{2}{c}{Predicted label}   \\ 
	\cline{2-3} 
	\multicolumn{1}{c}{Actual label}   & \multicolumn{1}{c}{Bad risk}  & \multicolumn{1}{c}{Good risk}   \\ 
	\midrule
	\multirow{2}{*}{Bad risk}    & $\pi_{0} F_{0}(\tau)$   &   $\pi_{0} (1 - F_{0}(\tau))$  \\ 
	& benefit: $0$                  &   cost: $B$                               \\ 
	\multirow{2}{*}{Good risk} & $\pi_{1} F_{1}(\tau)$     &    $\pi_{1} (1 - F_{1}(\tau))$   \\ 
	& cost: $C$                    &   benefit: $C$                               \\ 
	\bottomrule
\end{tabular} 
    \label{tab_costs}
\end{table}

% parameter B
The parameter $B$ reflects the cost associated with misclassifying a bad risk. Providing credit to a defaulter, the company faces a loss; specifically, the expected loss in case of default:
\begin{equation} 
B = \frac{\mbox{LGD} \cdot \mbox{EAD}}{A}\ ,
\end{equation}
where LGD refers to the loss given default, EAD is the exposure at default, and $A$ is the principal. $B$ varies between $0$ and $1$ and several distributions may arise \citep{somers2007quantile}. We follow \cite{RN36} and treat $B$ as a random variable with probability distribution:

\begin{itemize}
	\item $B = 0$ with probability $p_0$ (a customer repays the entire loan after default);
	\item $B = 1$ with probability $p_1$ (the bank loses the entire loan);
	\item $B$ follows a uniform distribution in (0, 1) with $F(B) = 1 - p_0 - p_1$.
\end{itemize}

% parameter C
The parameter $C$ reflects the opportunity cost or earned benefit associated with good risks. By accepting a good customer, the company earns the equivalent to the return on investment $\mbox{ROI}$: 
\begin{equation} 
C = \mbox{ROI} = \frac{I}{A}\ ,
\end{equation}
where $I$ is the total interest payments. Given these parameters, we compute profit as:
\begin{equation} 
\textnormal{Profit} = 
\int_{0}^{1} \Big[ C \cdot \big( \pi_{1} (1 - F_{1}  (\tau)) - \pi_{1} F_{1} (\tau) \big) - B \cdot \pi_{0} (1 - F_{0} (\tau)) \Big] 
f(B) d(B)
\end{equation}
% parameter values
This paper follows the empirical findings of \cite{RN36} and assumes a constant $\mbox{ROI}$ of $0.2664$ and the point masses $p_0 = 0.55$ for no loss and $p_1 = 0.1$ for full loss to compute $B$.

% performance evaluation
Apart from estimating the profitability of each fairness processor, we also compute the area under the ROC curve (AUC), which is a widely used indicator of the discriminatory ability of a scoring model. In addition, we evaluate fairness by measuring independence, separation and sufficiency. We aggregate the performance of pre- and post-processors over seven credit scoring data sets, five training fold combinations and four base classifiers, obtaining $140$ performance estimates per processor. Since in-processors do not require a base classifier, their performance is aggregated over $35$ values obtained from seven data sets and five training fold combinations.

\section{Empirical Results} 
\label{sec_6}

% intro
This section presents the empirical results. We first examine the correlation between the scorecard performance, profitability, and fairness. Next, we compare the performance of different fairness processors. Last, drawing on the findings that suggest a strong negative correlation between profit and fairness, we examine the profit-fairness trade-off to appraise the monetary cost of fairness.

%%%%%%%%%%%%%%%%%%%%%%%%%%%%
%% CORRELATION ANALYSIS
%%%%%%%%%%%%%%%%%%%%%%%%%%%%

\subsection{Correlation Analysis}

% results: intro + AUC vs Profit
Table \ref{tab_results_corr} depicts the mean Spearman correlation between the evaluation metrics. The correlation coefficients are computed on the performance estimates obtained from different variants of fairness processors and averaged over the seven credit scoring data sets. The results suggest that the AUC and profit often produce similar model rankings (correlation is $0.80$). Still, there is some disagreement between the two measures, which indicates that optimizing profit is important to identify potentially more profitable scorecards. Therefore, we emphasize profit in the following.

% results: fairness vs. profit
Comparing profit and fairness, we observe a moderate negative correlation between independence, separation, and profitability\footnote{Higher values of the AUC and profit indicate better performance, whereas lower values of independence, separation, and sufficiency indicate higher fairness. Therefore, we invert correlation signs between the two former performance metrics and the three fairness criteria to facilitate the consistent interpretation of the results.}. As expected, integrating fairness constraints to reduce discrimination prevents a scorecard from taking full advantage of the available information, which decreases profit. At the same time, a weak positive correlation between sufficiency and profit suggests that optimizing profitability without implementing additional fairness constraints could also improve sufficiency. This result confirms the observation that most ML algorithms are designed to automatically achieve sufficiency and implies that directly optimizing sufficiency with a fairness processor is not essential. 

% results: fairness criteria
A different conclusion emerges from examining the agreement of the other two fairness criteria. As indicated by Table \ref{tab_results_corr}, independence and separation have a strong positive correlation of $0.95$. Optimizing either of these two criteria will, therefore, favor models that fulfill both independence and separation. In other words, reducing the mutual information between a sensitive attribute and model predictions also helps to align the parity of error rates across the sensitive groups. This is an interesting finding, given that the former constraint targeted by independence is stricter compared to the one targeted by separation. For a risk analyst, the observed result implies that it is ample to rely on a single fairness criterion. Since separation has a better ability to capture the cost asymmetry (see Section \ref{sec_3} for details), we conclude that optimizing and measuring the separation criterion is the most suitable way to integrate and evaluate the fairness of a credit scoring model.

% mean correlation
\begin{table}[b]
    \centering
    \small
    \caption{Rank Correlation between Evaluation Metrics}
    \begin{tabular}{@{\extracolsep{10pt}} lccccc}
    \toprule
    Metric & AUC & Profit & IND & SP & SF \\
    \midrule 
    AUC    & 1          &            &           &           &  \\
    Profit & $0.8014$   & 1          &           &           &  \\
    IND    & $-0.4707$  & $-0.3774$  & 1         &           &  \\
    SP     & $-0.3326$  & $-0.2994$  & $0.9477$  & 1         &  \\
    SF     & $0.3489$   & $0.1636$   & $-0.2156$ & $-0.1311$ & 1 \\
    \bottomrule
    \multicolumn{6}{l}{Abbreviations: AUC = area under the ROC curve, IND = independence,} \\
    \multicolumn{6}{l}{SP = separation, SF = sufficiency.} \\
\end{tabular}

    \label{tab_results_corr}
\end{table}

%%%%%%%%%%%%%%%%%%%%%%%%%%%%
%% BENCHMARK RESULTS
%%%%%%%%%%%%%%%%%%%%%%%%%%%%

\subsection{Benchmarking Fairness Processors}

% table intro
Table \ref{tab_results_perf} provides average performance gains from fairness processors compared to the unconstrained scoring model across the seven credit scoring data sets. A positive gain indicates a better performance of a processor relative to the unconstrained model in terms of a particular evaluation measure. Individual results for each of the data sets are provided in the online Appendix.

% results: unconstrained model
Table \ref{tab_results_perf} confirms that using a processor to enhance fairness decreases profit compared to the unconstrained model. Results in terms of the AUC mirror this finding, whereby two processors show marginally higher AUC values than the unconstrained model. Table \ref{tab_results_perf} also evidences that the unconstrained model suffers from discrimination. Six out of eight processors achieve better independence and five processors attain better separation. However, sufficiency is consistently higher in the unconstrained model, which confirms that this metric differs fundamentally from independence and separation. High agreement between the sufficiency and profit, expressed by strict dominance of the unconstrained model in Table \ref{tab_results_perf}, also indicates that the goal of profit maximization is compatible with maximizing sufficiency, which questions the fairness perspective that the latter embodies.

% results: leaders
Considering individual processors, the reject option classification post-processor demonstrates the best fairness in independence and separation. This is achieved by sacrificing more than $30\%$ profit compared to the unconstrained model. On the other hand, we observe the least profit decrease of less than $5\%$ for the prejudice remover, which also attains a similar AUC as the unconstrained model. At the same time, the prejudice remover provides a smaller fairness improvement than other processors. These results emphasize the trade-off between profit and fairness.

% performance table
\begin{table}
    \centering
    \small
    \caption{Average Performance Gains from Fairness Processors Relative to the Unconstrained Model}
    \begin{tabular}{@{\extracolsep{5pt}} llccccc}
    \toprule
    Method & Fairness processor & AUC & Profit & IND & SP & SF \\ 
    \midrule 
    \multirow{2}{*}{Pre-processing} & Reweighting               & -3.19\%        & -23.04\%   & 66.00\%  & 61.24\%   & -38.18\%  \\
                                    & Disparate impact remover & \textbf{0.82}\% & -10.60\%  & 5.33\%  & 4.50\%   & \textbf{-19.99}\%  \\    
    \midrule
    \multirow{3}{*}{In-processing} & Prejudice remover     & 0.37\%    & \textbf{-4.28}\% & 11.51\%  & 9.41\%   & -202.36\%   \\
                                   & Adversarial debiasing & -0.21\%   & -13.90\%         & 9.38\%    & 2.98\%    & -148.36\%   \\    
                                   & Meta fair algorithm   & -2.98\%   & -7.25\%           & -7.49\%  & -20.88\% & -108.17\%   \\    
    \midrule
    \multirow{3}{*}{Post-processing} & Reject option classification & -8.64\%    & -30.71\%   & \textbf{74.80}\%  & \textbf{74.55}\%  & -263.51\%   \\
                                     & Equalized odds processor     & -16.22\%   & -59.73\%  & 25.83\%           & -11.08\%         & -407.82\%  \\     
                                     & Platt scaling                & -0.45\%   & -26.98\%  & -85.28\%          & -108.45\%         & -85.02\%  \\    
    \midrule
    \multicolumn{2}{l}{Average change across fairness processors} & -3.81\% & -22.06\% & 12.51\% & 1.53\% & -159.18\% \\
    \bottomrule
    \multicolumn{7}{l}{Abbreviations: AUC = area under the ROC curve, IND = independence, SP = separation, SF = sufficiency. Values} \\
    \multicolumn{7}{l}{represent percentage differences relative to an unconstrained scorecard averaged over seven data sets $\times$ five folds $\times$} \\
    \multicolumn{7}{l}{$\times$ four base models; positive values indicate improvement.} \\
\end{tabular}
    \label{tab_results_perf}
\end{table}

% results: best per group
Comparing processors within the implementation methods, we can identify promising techniques. Considering post-processors, the equalized odds processor is dominated by reject option classification in all evaluation measures. Platt scaling achieves higher profit and sufficiency than the latter but gives the by far worst results in independence and separation. In sum, Table \ref{tab_results_perf} clearly identifies reject option classification as the most suitable post-processor. Concerning pre-processors, no clear result emerges. Reweighing achieves the best fairness but decreases profitability by $23\%$. The disparate impact remover retains a red higher share of profit but offers substantially smaller improvements in independence and separation.

% results: in-processors
Among the in-processors, we observe the unconstrained model to dominate the meta fair algorithm, which displays negative results for all metrics of Table \ref{tab_results_perf}. Therefore, the meta fair algorithm does not warrant further consideration. Comparing the prejudice remover to adversarial debiasing, we find the former to deliver better results in all metrics but sufficiency. Given reservations against the fairness concept of the sufficiency metric, the results of Table \ref{tab_results_perf} suggest that the prejudice remover is the best performing in-processor. 

% implications I
The results of Table \ref{tab_results_perf} have several implications. First, we identify two fairness processors, Platt scaling and the meta-fair algorithm, inadequate for credit scoring since they decrease profit and predictive performance while not improving fairness compared to the unconstrained model. Second, we find that the equalized odds processor is dominated by another post-processor in all considered evaluation metrics and should, therefore, be avoided.

% implications II
The remaining processors arrive at different solutions in the space between sacrificing profit and reducing discrimination, leaving decision-makers with the difficult task to balance these conflicting goals according to their preferences, business requirements, and regulation. In general, in-processors offer more flexibility in prioritizing fairness or profit through meta-parameters. For example, the prejudice remover incorporates a regularizer to penalize fairness violations and exposes the weight of that penalty as a meta-parameter. However, the benefit of higher flexibility carries a cost. Compared to alternative options, in-processors replace existing scorecards and impact the scoring process the most. Post-processors largely retain an existing scoring pipeline, which simplifies their deployment. Pre-processors address fairness at the data level, which represents a more invasive change of the scoring process compared to post-processing but seems less difficult to implement than in-processing. Together with the results of Table \ref{tab_results_perf}, in which the best in-processor (i.e., the prejudice remover) finds a better trade-off between profit and fairness than the disparate impact remover while the best post-processor (i.e., reject option classification) increases fairness to a larger extent than reweighting, considerations related to the complexity of deploying fairness processors and revising loan approval processes suggest two options for addressing fairness in credit scoring. Decision-makers can choose between a flexible but invasive in-processor and a post-processor, which is easier to deploy but might substantially decrease profitability. Table \ref{tab_results_perf} represents the corresponding options by the prejudice remover and reject option classification.

%%%%%%%%%%%%%%%%%%%%%%%%%%%%
%% PARETO FRONTIERS
%%%%%%%%%%%%%%%%%%%%%%%%%%%%

\subsection{The Cost of Fairness}

\begin{figure}[b]
	\centering
    \includegraphics[width = \linewidth]{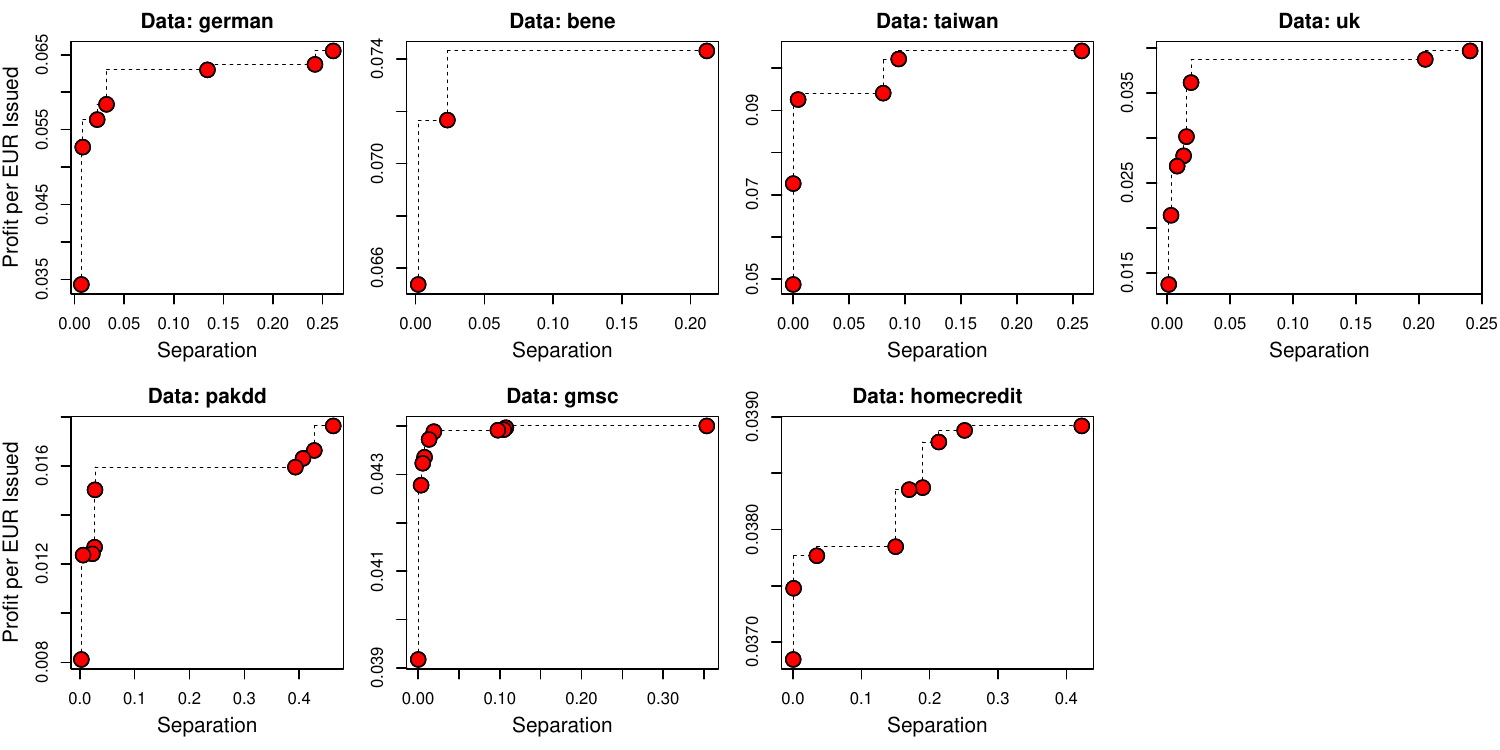}
    \caption{Profit-Fairness Trade-Off: Frontiers with Non-Dominated Solutions}
	\label{fig_profit_fairness}
\end{figure}

% frontiers description
Previous results indicate that it is possible to improve fairness by sacrificing profit. Figure \ref{fig_profit_fairness} provides a more detailed examination of the profit-fairness trade-off on each of the seven data sets using the concept of Pareto frontiers. The points on the frontiers refer to the test set performance of fairness processors trained with different base classifiers and on different combinations of the training folds. The frontiers only contain the non-dominated solutions, i.e., the points where it is impossible to improve on one objective (i.e., profit) without harming the other objective (i.e., fairness). Based on the previous results, we use the separation criterion to measure fairness.

% results discussion
Figure \ref{fig_profit_fairness} reveals that discrimination can be substantially reduced at a relatively low cost. Recall that separation indicates the difference between the false positive and false negative rates across the sensitive groups. According to Figure \ref{fig_profit_fairness}, reducing the difference in error rates below $0.2$ is possible while sacrificing less than \euro{0.01} profit per EUR issued. Across the data sets, this translates to an average profit reduction of $4.91\%$ compared to the most profitable scorecard with stronger discrimination. At the same time, completely eliminating unfairness is more costly: achieving separation of $0$ is only possible when sacrificing more than $35\%$ of the profit. However, since perfect fairness is not required by regulation, we conclude that a financial institution can reduce discrimination to a reasonable extent while maintaining a relatively high profit margin.

% profit vs separation

\section{Conclusion} 
\label{sec_conclusion}

% summary
The paper sets out to consolidate recent advancements in fair ML from a credit scoring perspective. Cataloging approaches for quantifying fairness and the ML pipeline interventions for fairness maximization, we have examined the adequacy of these fairness measures and processors for credit scoring. To substantiate our conceptual analysis, we have undertaken a systematic empirical comparison of several fairness processors from different families to identify preferable approaches and clarify the degree to which increasing fairness in loan approval processes harms profitability.

% catalog and criteria
The conceptual comparison of different fairness criteria reveals separation to be the most appropriate metric for credit scoring. Separation acknowledges the imbalanced misclassification costs, which are instrumental to the lending business. The presented catalog of fairness processors offers practitioners a starting point for deciding which processors to consider for a given problem setting. The catalog also indicates that most processors have been evaluated based on their accuracy and that some relevant credit scoring scenarios are not well covered by the available processors. For example, in a setting with multiple sensitive attributes (e.g., race and religion), only two processors, adversarial debiasing and reject option classification, facilitate optimizing the separation criterion. 

% fairness criteria 
The empirical study benchmarks fairness processors in a profit-oriented credit scoring setup. Several implications emerge from the results. First, examining the agreement between the fairness criteria under study reveals that separation and independence are strongly correlated. While other empirical studies support this finding \citep{RN11}, it contradicts the intuition from theoretical considerations that fairness criteria are mutually exclusive \citep{RN8}. We also find that sufficiency has a property to be achievable by any well-trained classifier that can predict the sensitive attribute from the other features \citep{RN7}. This calls into question the overall suitability of sufficiency for credit scoring and further emphasizes separation as a proper criterion for measuring the fairness of credit scorecards.

% fairness processors
Second, we find that the choice of an appropriate fairness processor depends on the implementation feasibility and preferences of a decision-maker regarding the conflicting objectives of profit and fairness. Post-processing methods such as reject option classification are the easiest to implement in production but improve fairness at a high monetary cost. In-processors such as the prejudice remover perform best in finding the profit-fairness trade-off and offer the most flexibility in calibrating the importance of the conflicting objectives. However, using in-processors requires replacing a deployed scoring model with a new algorithm, which might require regulatory approval and is, more generally, associated with considerable efforts.

% profit-fairness trade-off
Third, while achieving perfect fairness is costly, we find that reducing discrimination to a reasonable extent is possible while maintaining a relatively high profit. These results support the current anti-discrimination regulation that allows unfairness to exist up to a certain limited extent. The analysis of fairness processors from the perspective of the Pareto frontiers offers decision-makers a tool to analyze the profit-fairness trade-off specific to their context and identify techniques that reduce discrimination to a required level at the smallest monetary cost.

% other domains
Our study may also have implications for customer scoring models beyond the credit industry. Fairness concerns arise from the increasing use of ML to automate decisions in many domains, such as hiring \citep{RN7}, college admission \citep{RN8} or criminal risk assessment \citep{RN43}. The catalog of fairness processors and the results of their empirical analysis can aid these domains in identifying suitable techniques for integrating fairness in decision support systems. Future work on fair ML may also draw value from the empirical comparison in that it highlights effective approaches that set a benchmark for new fairness processors.

%%%%%%%%%%%%%%%%%%%%%%%%%%%%%%%%%%%%
%                                  
%            REFERENCES
%                                  
%%%%%%%%%%%%%%%%%%%%%%%%%%%%%%%%%%%%


%\bibliographystyle{model5-names}
%\biboptions{authoryear}
%\bibliography{references.bib}


%%%%%%%%%%%%%%%%%%%%%%%%%%%%%%%%%%%%
%                                  
%           APPENDIX
%                                  
%%%%%%%%%%%%%%%%%%%%%%%%%%%%%%%%%%%%


\newpage

\appendix

\section*{Online Appendix}

\section{Overview of Fairness Criteria} 
\label{sec_a1}

% intro
As shown by \citet{RN7}, the fairness criteria considered in this paper -- independence, separation and sufficiency -- comprise a number of other fairness criteria proposed in prior work. This appendix illustrates the relationship between the three criteria and related fairness formulations.

% Table 1 conclusions
Table \ref{tab_criteria} reveals that the statistical formulation of fairness constraints originates from the field of psychological testing \citep{RN42} and has been rediscovered for ML applications much later. The 19 fairness concepts presented in the table can be derived from independence, separation and sufficiency in the form of an equivalent or a relaxed condition. This underpins the relevance of the three fairness criteria and justifies our criteria selection in the focal paper. 

% Acknowledge fairness concepts beyond group-based fairness
However, it is important to note that all fairness criteria of Table \ref{tab_criteria} and the paper as a whole embody the idea of group-based fairness. Prior literature has introduced alternative fairness concepts including individual and counterfactual fairness. The former requires a classifier to produce similar outputs for similar individuals, whereas the latter implies that a classifier output remains the same when the sensitive attribute is changed to its counterfactual value.

% fairness criteria
\begin{table}[H]
    \centering
    \small
    \caption{Fairness Criteria and their Relation to Independence, Separation, and Sufficiency}
    \begin{tabular}{llll}
    \toprule
    Reference & Criterion & Closest relative & Relation  \\ 
    \midrule
    \cite {RN42} & Darlington criterion (4) & Independence & Equivalent  \\
    \cite {RN23} & Statistical parity & Independence & Equivalent  \\
    \cite {RN23} & Group fairness & Independence & Equivalent  \\
    \cite {RN23} & Demographic parity & Independence & Equivalent  \\
    \cite {RN41} & Conditional statistical parity & Independence & Relaxation  \\
    \midrule
    \cite {RN42} & Darlington criterion (3) & Separation & Relaxation  \\
    \cite {RN18} & Equal opportunity & Separation & Relaxation  \\
    \cite {RN18} & Equalized odds & Separation & Equivalent  \\
    \cite {RN21} & Balance for the negative class & Separation & Relaxation  \\
    \cite {RN21} & Balance for the positive class & Separation & Relaxation  \\
    \cite {RN30} & Avoiding disparate mistreatment & Separation & Equivalent  \\
    \cite {RN20} & Predictive equality & Separation & Relaxation  \\
    \cite {RN45} & Equalized correlations & Separation & Relaxation  \\
    \cite {RN43} & Conditional procedure accuracy & Separation & Equivalent  \\
    \midrule
    \cite {RN44} & Cleary model & Sufficiency & Equivalent  \\
    \cite {RN42} & Darlington criterion (1), (2) & Sufficiency & Relaxation  \\
    \cite {RN20} & Predictive parity & Sufficiency & Relaxation  \\
    \cite {RN20} & Calibration within groups & Sufficiency & Equivalent  \\
    \cite {RN43} & Conditional use accuracy & Sufficiency & Equivalent  \\
    \bottomrule
\end{tabular}
    \label{tab_criteria}
\end{table}

\vspace{1.5cm}
\section{Meta-Parameters of Base Models and Fairness Processors} 
\label{sec_a2}

% description: base models
This appendix provides meta-parameter values of the base classifiers and the fairness processors used in the empirical experiment. Table \ref{tab_params_base} depicts the candidate values of the meta-parameters of the four base classifiers used as a scoring model by fairness pre- and post-processors as well as by the unconstrained profit maximization benchmark. The meta-parameter values are optimized with grid search using the EMP as an objective. The meta-parameter tuning is performed separately on each combination of the training folds using a nested four-fold cross-validation.

% description: fairness processors
Table \ref{tab_params_fair} provides candidate values of the meta-parameters of fairness processors that are tuned within the higher-level cross-validation framework. We measure the EMP of fairness processors on each validation fold to select the appropriate meta-parameter values. The notation for processor meta-parameters and their explanation is available in Section 4.

% params: processors
\begin{table}[H]
    \centering
    \small
    \caption{Meta-Parameters of Fairness Processors}
    \begin{tabular}{llll}
    \toprule
    Method & Fairness processor & Meta-parameter & Candidate values \\
    \midrule
    \multirow{2}{*}{Pre-processing} & Reweighting & -- & -- \\
                         \cmidrule{2-4}
                         & Disparate impact remover & Repair level $\lambda$ & $0.5, 0.6, 0.7, 0.8, 0.9, 1$ \\
    \midrule
    \multirow{5}{*}{In-processing} & Prejudice remover & Fairness penalty $\eta$ & $1, 5, 15, 30, 50, 70, 100, 150$ \\
                        \cmidrule{2-4}
                        & Meta fair algorithm & Fairness penalty $\tau$ & $0.05, 0.10, 0.15, 0.20, 0.25, 0.30$ \\
                        \cmidrule{2-4}
                        & \multirow{3}{*}{Adversarial debiasing} & Adversarial loss weight $\alpha$ & $0.1, 0.01, 0.001$ \\
                        &                                        & Number of epochs                 & $50$ \\
                        &                                        & Batch size                       & $128$ \\
    \midrule
    \multirow{5}{*}{Post-processing} & \multirow{3}{*}{Reject option classification} & Fairness bound $[\sigma_l, \sigma_u]$ & $[-0.1, 0.1], [-0.2, 0.2], [-0.3, 0.3]$ \\
                          &                                               & Number of thresholds     & $100$ \\
                          &                                               & Number of ROC margins    & $50$ \\ 
                          \cmidrule{2-4}
                          & \multirow{1}{*}{Equalized odds processor}     & -- & -- \\
                          \cmidrule{2-4}
                          & \multirow{1}{*}{Platt scaling}                & -- & -- \\
    \bottomrule
\end{tabular}
    \label{tab_params_fair}
\end{table}

% params: base models
\begin{table}[H]
    \centering
    \small
    \caption{Meta-Parameters of Base Classifiers}
    \begin{tabular}{llll}
    \toprule
    Base classifier & Meta-parameter & Candidate values \\
    \midrule
    Logistic regression & -- & -- \\
    \midrule
    \multirow{2}{*}{Random forest} & Number of trees            & $500$       \\
                                   & Number of sampled features & $5, 10, 15$ \\
    \midrule
    \multirow{6}{*}{Extreme gradient boosting} & Number of trees            & $100, 500, 1000$ \\
                                       & Maximum tree depth         & $5, 10$          \\    
                                       & Learning rate              & $0.1$            \\                                   
                                       & Ratio  of sampled features & $0.5, 1$         \\                                   
                                       & Ratio  of sampled cases    & $0.5, 1$         \\                                
                                       & Minimum child weight       & $0.5, 1, 3$      \\
    \midrule
    \multirow{3}{*}{Artificial neural network} & Size                        & $5, 10, 15$           \\
                                    & Decay                       & $0.1, 0.5, 1, 1.5, 2$ \\
                                    & Maximum umber of iterations & $1000$                \\
    \bottomrule
\end{tabular}
    \label{tab_params_base}
\end{table}

\section{Extended Empirical Results} 
\label{sec_a3}

% summary: tables
This appendix provides additional results of the experiment presented in Section 6. Tables \ref{tab_res_german} -- \ref{tab_res_homecredit} compare the performance of fairness processors as well as an unconstrained scorecard on each of the seven credit scoring data sets in terms of the AUC, profit per EUR issued and fairness. Performance of pre- and post-processors is averaged over $25$ values from five cross-validation folds $\times$ five base classifiers; performance of in-processors is aggregated over five training fold combinations.

% individual results: german
\begin{table}[H]
    \centering
    \small
    \caption{Performance of Fairness Processors: German}
    \begin{tabular}{@{\extracolsep{8pt}} llcccccc}
    \toprule
    Method & Fairness processor & AUC & Profit & AR & IND & SP & SF \\
    \midrule 
    \multirow{2}{*}{Pre-processing} & Reweighing               & .7604  & .0252  & .6113  & .2204  & .1752 & .1563 \\
                         & Disparate impact remover & .8121  & .0494  & .6172  & .2989  & .1919 & .1249 \\
    \midrule
    \multirow{3}{*}{In-processing} & Prejudice remover     & .7933  & .0463  & .6112  & .3200  & .2091 & .1655 \\
                        & Adversarial debiasing & .7965  & .0502  & .6103  & .2528  & .1898 & .1705 \\
                        & Meta fair algorithm   & .8074  & .0467  & .6158  & .2262  & .1117 & .1555 \\
    \midrule
    \multirow{3}{*}{Post-processing} & Reject option classification & .7124  & .0254  & .5985  & .1105  & .0881 & .2121 \\
                          & Equalized odds processor     & .6999  & .0300  & .5965  & .0836  & .1475 & .2514 \\
                          & Platt scaling                & .8012  & .0464  & .6139  & .4195  & .3369 & .1532 \\
    \midrule
    \multicolumn{2}{l}{Unconstrained profit maximization} & .8124  & .0492  & .6143  & .3078  & .1979 & .1445 \\
    \bottomrule
    \multicolumn{8}{l}{Abbreviations: AUC = area under the ROC curve, AR = acceptance rate, IND = independence, SP = separation,} \\
    \multicolumn{8}{l}{SF = sufficiency. Performance is averaged over five cross-validation folds $\times$ four base models.} \\
\end{tabular}
    \label{tab_res_german}
\end{table}

% individual results: bene
\begin{table}[H]
    \centering
    \small
    \caption{Performance of Fairness Processors: Bene}
    \begin{tabular}{@{\extracolsep{8pt}} llcccccc}
    \toprule
    Method & Fairness processor & AUC & Profit & AR & IND & SP & SF \\
    \midrule 
    \multirow{2}{*}{Pre-processing} & Reweighing               & .7469 & .0524  & .6108  & .0934  & .0777 & .0487 \\
                                    & Disparate impact remover & .7875 & .0638  & .6138  & .3622  & .2832 & .0694 \\
    \midrule
    \multirow{3}{*}{In-processing} & Prejudice remover     & .7952  & .0702  & .6194  & .3141  & .2284 & .1615 \\
                                  & Adversarial debiasing & .7813  & .0670  & .6118  & .3250  & .2450 & .1393 \\
                                  & Meta fair algorithm   & .7875  & .0653  & .6143  & .3227  & .2446 & .0980 \\
    \midrule
    \multirow{3}{*}{Post-processing} & Reject option classification & .7082  & .0501  & .6037  & .0726  & .0711 & .2339 \\
                                     & Equalized odds processor     & .6677  & .0491  & .6039  & .0396  & .0844 & .2485 \\
                                     & Platt scaling                & .7880  & .0659  & .6158  & .4469  & .3684 & .0681 \\  
    \midrule
    \multicolumn{2}{l}{Unconstrained profit maximization} & .7896  & .0659  & .6152  & .3540  & .2743 & .0825 \\
    \bottomrule
    \multicolumn{8}{l}{Abbreviations: AUC = area under the ROC curve, AR = acceptance rate, IND = independence, SP = separation,} \\
    \multicolumn{8}{l}{SF = sufficiency. Performance is averaged over five cross-validation folds $\times$ four base models.} \\
\end{tabular}
    \label{tab_res_bene}
\end{table}

% individual results: taiwan
\begin{table}
    \centering
    \small
    \caption{Performance of Fairness Processors: Taiwan}
    \begin{tabular}{@{\extracolsep{8pt}} llcccccc}
    \toprule
    Method & Fairness processor & AUC & Profit & AR & IND & SP & SF \\
    \midrule 
    \multirow{2}{*}{Pre-processing} & Reweighing               & .7605  & .0725  & .5954  & .0441  & .0406 & .0189 \\
                                    & Disparate impact remover & .7909  & .0769  & .5999  & .1781  & .1324 & .0216 \\
    \midrule
    \multirow{3}{*}{In-processing} & Prejudice remover     & .7867  & .0882  & .5992  & .1170  & .0876 & .0311 \\
                                   & Adversarial debiasing & .7918  & .0892  & .5987  & .2762  & .2262 & .0215 \\
                                   & Meta fair algorithm   & .7893  & .0880  & .6009  & .1212  & .0854 & .0197 \\
    \midrule
    \multirow{3}{*}{Post-processing} & Reject option classification & .7080  & .0515  & .5869  & .0514  & .0328 & .0505 \\
                                     & Equalized odds processor     & .6231  & -.0081 & .5797  & .1902  & .1915 & .0618 \\
                                     & Platt scaling                & .7565  & .0294  & .5960  & .2738  & .2187 & .0159 \\  
    \midrule
    \multicolumn{2}{l}{Unconstrained profit maximization} & .7532  & .0643  & .5956  & .1211  & .0872 & .0278 \\
    \bottomrule
    \multicolumn{8}{l}{Abbreviations: AUC = area under the ROC curve, AR = acceptance rate, IND = independence, SP = separation,} \\
    \multicolumn{8}{l}{SF = sufficiency. Performance is averaged over five cross-validation folds $\times$ four base models.} \\
\end{tabular}
    \label{tab_res_taiwanese}
\end{table}

% individual results: uk
\begin{table}
    \centering
    \small
    \caption{Performance of Fairness Processors: UK}
    \begin{tabular}{@{\extracolsep{8pt}} llcccccc}
    \toprule
    Method & Fairness processor & AUC & Profit & AR & IND & SP & SF \\
    \midrule 
    \multirow{2}{*}{Pre-processing} & Reweighing               & .6786  & .0165  & .5544  & .0807  & .0396 & .0056 \\
                         & Disparate impact remover & .7174  & .0131  & .5543  & .2926  & .2051 & .0113 \\
    \midrule
    \multirow{3}{*}{In-processing} & Prejudice remover     & .7087  & .0187  & .5543  & .3033  & .2433 & .0191 \\
                        & Adversarial debiasing & .7092  & .0181  & .5543  & .2614  & .1622 & .0181 \\
                        & Meta fair algorithm   & .5584  & .0038  & .5542  & .3389  & .4128 & .0021 \\
    \midrule
    \multirow{3}{*}{Post-processing} & Reject option classification & .6523  & .0181  & .5542  & .0621  & .0222 & .0162 \\
                          & Equalized odds processor     & .6206  & .0189  & .5542  & .1783  & .2042 & .0225 \\
                          & Platt scaling                & .6986  & .0244  & .5543  & .6839  & .5329 & .0200 \\  
    \midrule
    \multicolumn{2}{l}{Unconstrained profit maximization} & .7129  & .0180  & .5543  & .3111  & .2141 & .0141 \\
    \bottomrule
    \multicolumn{8}{l}{Abbreviations: AUC = area under the ROC curve, AR = acceptance rate, IND = independence, SP = separation,} \\
    \multicolumn{8}{l}{SF = sufficiency. Performance is averaged over five cross-validation folds $\times$ four base models.} \\
\end{tabular}
    \label{tab_res_uk}
\end{table}

% individual results: pakdd
\begin{table}
    \centering
    \small
    \caption{Performance of Fairness Processors: PAKDD}
    \begin{tabular}{@{\extracolsep{8pt}} llcccccc}
    \toprule
    Method & Fairness processor & AUC & Profit & AR & IND & SP & SF \\
    \midrule 
    \multirow{2}{*}{Pre-processing} & Reweighing               & .5783  & .0078  & .5836  & .0710  & .0685 & .0198 \\
                         & Disparate impact remover & .6022  & .0134  & .5840  & .4126  & .3862 & .0818 \\
    \midrule
    \multirow{3}{*}{In-processing} & Prejudice remover     & .6003  & .0134  & .5839  & .2829  & .2506 & .1171 \\
                        & Adversarial debiasing & .5777  & .0079  & .5835  & .1864  & .1686 & .0952 \\
                        & Meta fair algorithm   & .6027  & .0136  & .5839  & .3383  & .3070 & .1138 \\
    \midrule
    \multirow{3}{*}{Post-processing} & Reject option classification & .5677  & .0080  & .5834  & .0822  & .0602 & .1153 \\
                          & Equalized odds processor     & .5653  & .0112  & .5833  & .0173  & .0394 & .1214 \\
                          & Platt scaling                & .6053  & .0144  & .5840  & .6249  & .5920 & .1044 \\  
    \midrule
    \multicolumn{2}{l}{Unconstrained profit maximization} & .6045  & .0139  & .5840  & .4347 & .4069 & .0829 \\
    \bottomrule
    \multicolumn{8}{l}{Abbreviations: AUC = area under the ROC curve, AR = acceptance rate, IND = independence, SP = separation,} \\
    \multicolumn{8}{l}{SF = sufficiency. Performance is averaged over five cross-validation folds $\times$ four base models.} \\
\end{tabular}
    \label{tab_res_pakdd}
\end{table}

% individual results: gmsc
\begin{table}
    \centering
    \small
    \caption{Performance of Fairness Processors: GMSC}
    \begin{tabular}{@{\extracolsep{8pt}} llcccccc}
    \toprule
    Method & Fairness processor & AUC & Profit & AR & IND & SP & SF \\
    \midrule 
    \multirow{2}{*}{Pre-processing} & Reweighing               & .8425  & .0415  & .5589  & .0595  & .0437 & .0055  \\
                         & Disparate impact remover & .8535  & .0419  & .5593  & .1935  & .1077 & .0151  \\
    \midrule
    \multirow{3}{*}{In-processing} & Prejudice remover     & .8553  & .0437  & .5593  & .2454  & .1445 & .0085 \\
                        & Adversarial debiasing & .8588  & .0438  & .5594  & .1126  & .0508 & .0154 \\
                        & Meta fair algorithm   & .8261  & .0418  & .5593  & .4318  & .2766 & .0182 \\
    \midrule
    \multirow{3}{*}{Post-processing} & Reject option classification & .7762  & .0388  & .5576  & .0590  & .0525 & .0187 \\
                          & Equalized odds processor     & .5903  & .0234  & .5568  & .3844  & .2812 & .0240 \\
                          & Platt scaling                & .8531  & .0424  & .5594  & .5691  & .3564 & .0000 \\  
    \midrule
    \multicolumn{2}{l}{Unconstrained profit maximization} & .8545  & .0406  & .5594  & .2461  & .1429 & .0121 \\
    \bottomrule
    \multicolumn{8}{l}{Abbreviations: AUC = area under the ROC curve, AR = acceptance rate, IND = independence, SP = separation,} \\
    \multicolumn{8}{l}{SF = sufficiency. Performance is averaged over five cross-validation folds $\times$ four base models.} \\
\end{tabular}
    \label{tab_res_gmsc}
\end{table}

% individual results: homecredit
\begin{table}
    \small
    \caption{Performance of Fairness Processors: Homecredit}
    \begin{tabular}{@{\extracolsep{8pt}} llcccccc}
    \toprule
    Method & Fairness processor & AUC & Profit & AR & IND & SP & SF \\
    \midrule 
    \multirow{2}{*}{Pre-processing} & Reweighing               & .7275  & .0353  & .5589  & .1225  & .0958 & .0056 \\
                         & Disparate impact remover & .7392  & .0361  & .5590  & .2784  & .2010 & .0167 \\
    \midrule
    \multirow{3}{*}{In-processing} & Prejudice remover     & .7387  & .0372  & .5589  & .2072  & .1356 & .0238 \\
                        & Adversarial debiasing & .7379  & .0371  & .5589  & .3170  & .2464 & .0169 \\
                        & Meta fair algorithm   & .7351  & .0367  & .5588  & .3482  & .2661 & .0058 \\
    \midrule
    \multirow{3}{*}{Post-processing} & Reject option classification & .6785  & .0350  & .5585  & .0506  & .0207 & .0290 \\
                          & Equalized odds processor     & .6190  & .0260  & .5583  & .2481  & .2482 & .0426 \\
                          & Platt scaling                & .7406  & .0371  & .5590  & .4884  & .3643 & .0140 \\  
    \midrule
    \multicolumn{2}{l}{Unconstrained profit maximization} & .7411  & .0367  & .5590  & .33044 & .2435 & .0130 \\
    \bottomrule
    \multicolumn{8}{l}{Abbreviations: AUC = area under the ROC curve, AR = acceptance rate, IND = independence, SP = separation,} \\
    \multicolumn{8}{l}{SF = sufficiency. Performance is averaged over five cross-validation folds $\times$ four base models.} \\
\end{tabular}
    \label{tab_res_homecredit}
\end{table}

\newpage
\bibliographystyle{model5-names}

\end{document}